%% file: main.tex
\documentclass[10pt,twocolumn,letterpaper]{article}

\usepackage{cvpr}      
\usepackage{graphicx}
\usepackage{fontawesome5} 
\input{preamble}
\definecolor{cvprblue}{rgb}{0.21,0.49,0.74}
\usepackage[pagebackref,breaklinks,colorlinks,allcolors=cvprblue]{hyperref}
\def\paperID{8338} 
\def\confName{CVPR}
\def\confYear{2026}

\title{EventBench: Towards Comprehensive Benchmarking of Event-based MLLMs}

\author{Shaoyu Liu$^{1, 2}$, Jianing Li$^{2}$, Guanghui Zhao$^{1}$, Yunjian Zhang$^{2}$, Xiangyang Ji$^{2}$\\
\\
$^{1}$Xidian University ~ $^{2}$Tsinghua University  \\
\\
\\
\faGithub\;\textbf{Project Page \& Code}: \href{https://xdusyl.github.io/eventbench.github.io/}{\textcolor{cvprblue}{eventbench.github.io}}
\quad\quad
\includegraphics[height=1.0em]{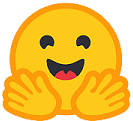}\;
\textbf{Dataset:}\;
\href{https://huggingface.co/datasets/XduSyL/EevntBench}{\textcolor{cvprblue}{eventbench.huggingface}}
}

\begin{document}
\maketitle
\input{sec/0_abstract}

\input{sec/1_intro}
\input{sec/2_related_works}

\input{sec/3_method}

\input{sec/4_experiment}
\input{sec/5_conclusion}
{
    \small
    \bibliographystyle{ieeenat_fullname}
    \bibliography{main}
}
\input{sec/X_suppl}


\end{document}

%% file: sec/0_abstract.tex
\begin{abstract}
Multimodal large language models (MLLMs) have made significant advancements in event-based vision. 
However, the comprehensive evaluation of these models' capabilities within a unified benchmark remains a crucial yet largely unexplored aspect.
In this paper, we introduce \textbf{EventBench}, which presents an evaluation benchmark covering 8 diverse task metrics and a large-scale event stream dataset.
Our work distinguishes existing event-based benchmarks in four key aspects:
\textbf{i) Openness in accessibility}, releasing all raw event streams and task instructions across 8 evaluation metrics;
\textbf{ii) Diversity in task coverage}, covering diverse understanding, recognition, and spatial reasoning tasks, enabling comprehensive evaluation of model capability;
\textbf{iii) Integration in spatial dimensions}, pioneering the design of 3D spatial reasoning task for event-based MLLMs;
\textbf{iv) Scale in data volume}, with an accompanying training set of over one million event–text pairs supporting large-scale training and evaluation.
With EventBench, we evaluate state-of-the-art closed-source models such as GPT-5 and Gemini-2.5 Pro, leading open-source models including Qwen2.5-VL and InternVL3, as well as event-based MLLMs like EventGPT that directly process raw event inputs. Extensive evaluation reveals that current event-based MLLMs perform well in event stream understanding, yet they still struggle with fine-grained recognition and spatial reasoning.

\end{abstract}

%% file: sec/1_intro.tex
\section{Introduction}
\label{sec:intro}

\begin{figure}[t]
  \centering
   \includegraphics[width=\linewidth]{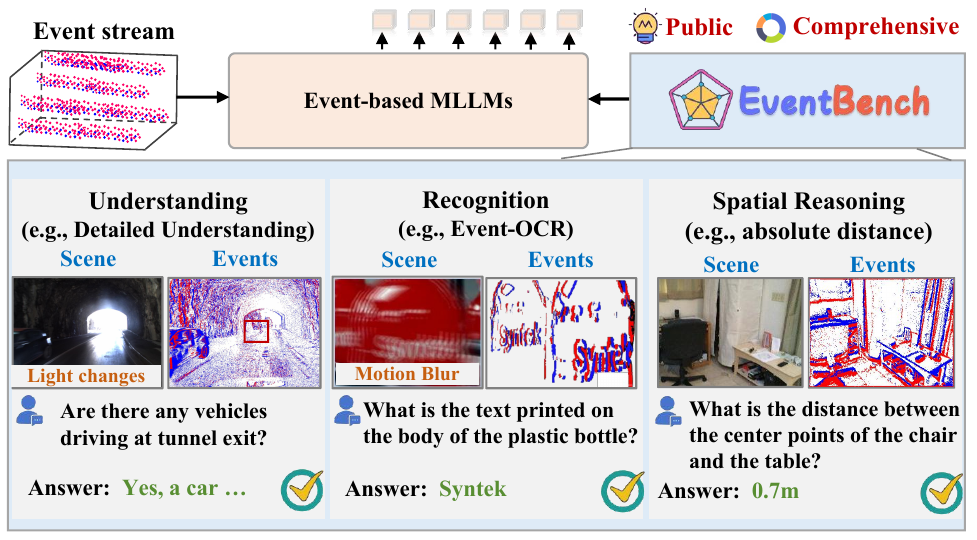}
   \caption{Our EventBench is a publicly accessible and comprehensive evaluation benchmark for event-based MLLMs. It offers diverse task metrics across multiple dimensions (i.e., understanding, recognition, and spatial reasoning) and will accelerate research on event-based MLLMs in challenging scenarios.}
   \vspace{-10pt}
   \label{fig:motivation}
\end{figure}

\begin{table*}[t]
\centering
\caption{Comparison of event-based MLLM datasets and benchmarks. 
We compare datasets in terms of scale, modalities, evaluation metrics, and four key attributes:
\textcolor[rgb]{0.129, 0.298, 0.494}{$\mathbf{\Delta d}$} = Downstream task support, 
\textcolor[rgb]{0.149, 0.361, 0.259}{$\mathbf{\Delta s}$} = Spatial task support, 
\textcolor[rgb]{0.471, 0.141, 0.196}{$\mathbf{\Delta t}$} = Training data release, 
\textcolor[rgb]{0.600, 0.298, 0.110}{$\mathbf{\Delta b}$} = Benchmark release.
Time ranges are categorized as \emph{short} (50\,ms), \emph{medium} (50\,ms--10\,s), and \emph{long} ($>$10\,s).
(\protect\includegraphics[height=0.9em]{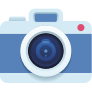} RGB,\;
\protect\includegraphics[height=0.9em]{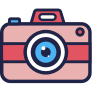} Event Stream, and \;
\protect\includegraphics[height=0.9em]{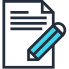} Text)}
\renewcommand{\arraystretch}{1.2}
\resizebox{\textwidth}{!}{
\begin{tabular}{l |c c c c c c | c c c c}
\toprule
\multirow{2}{*}{\textbf{Dataset}} &
\multirow{2}{*}{\textbf{Venue}} &
\multirow{2}{*}{\textbf{Size}} &
\multirow{2}{*}{\textbf{Modalities}} &
\multirow{2}{*}{\textbf{Metric Types}} &
\multirow{2}{*}{\textbf{Data Collection Source}} &
\multirow{2}{*}{\textbf{Temporal Span}} &
\multicolumn{4}{c}{\textbf{Attributes}} \\
\cmidrule(lr){8-11}
& & & & & & & \textcolor[rgb]{0.129, 0.298, 0.494}{$\mathbf{\Delta d}$} & \textcolor[rgb]{0.149, 0.361, 0.259}{$\mathbf{\Delta s}$} & \textcolor[rgb]{0.471, 0.141, 0.196}{$\mathbf{\Delta t}$} & \textcolor[rgb]{0.600, 0.298, 0.110}{$\mathbf{\Delta b}$} \\
\midrule
\textbf{N-ImageNet}\cite{kim2021n} & ICCV'21 & 1M &
\includegraphics[height=0.9em]{images/event_camera_logo.png}
& -- & Synthesis & Short & \xmark & \xmark & \cmark & \cmark \\
\textbf{Talk2Event}\cite{kong2025talk2event} & NeurIPS'25 & 30.7K &
\includegraphics[height=0.9em]{images/RGB_logo.png}\,\includegraphics[height=0.9em]{images/event_camera_logo.png}\,\includegraphics[height=0.9em]{images/text_logo.png}
& -- & Real World & Short & \cmark & \xmark & \xmark & \xmark \\
\textbf{EventChat}\cite{liu2025eventgpt} & CVPR'25 & 1.1M &
\includegraphics[height=0.9em]{images/event_camera_logo.png}\,\includegraphics[height=0.9em]{images/text_logo.png}
& 3 & Synthesis \& Real World & Short & \xmark & \xmark & \cmark & \xmark \\
\textbf{EventVL-Bench}\cite{li2025eventvl} & arXiv'25 & 1.4M &
\includegraphics[height=0.9em]{images/RGB_logo.png}\,\includegraphics[height=0.9em]{images/event_camera_logo.png}\,\includegraphics[height=0.9em]{images/text_logo.png}
& 4 & Synthesis \& Real World & Short \& Medium & \cmark & \xmark & \xmark & \xmark \\
\textbf{LLaFEA-Bench}\cite{zhou2025llafea} & ICCV'25 & -- &
\includegraphics[height=0.9em]{images/RGB_logo.png}\,\includegraphics[height=0.9em]{images/event_camera_logo.png}\,\includegraphics[height=0.9em]{images/text_logo.png}
& 3 & Synthesis \& Real World & Short \& Medium & \cmark & \xmark & \xmark & \xmark \\
\textbf{EVQA-Bench}\cite{chen2025let} & arXiv'25 & 1.0M &
\includegraphics[height=0.9em]{images/event_camera_logo.png}\,\includegraphics[height=0.9em]{images/text_logo.png}
& 4 & Synthesis \& Real World & Short \& Medium \& Long & \xmark & \xmark & \xmark & \xmark \\
\midrule
\rowcolor{gray!20}
\textbf{EventBench} & \textbf{Ours*} & \textbf{1.4M} &
\includegraphics[height=0.9em]{images/RGB_logo.png}\,\includegraphics[height=0.9em]{images/event_camera_logo.png}\,\includegraphics[height=0.9em]{images/text_logo.png}
& \textbf{8} & \textbf{Synthesis \& Real World} & \textbf{Short \& Medium \& Long} & \cmark & \cmark & \cmark & \cmark \\
\bottomrule
\end{tabular}
}
\label{tab:datasets}
\end{table*}

Event cameras, offering high temporal resolution and high dynamic range, have been widely applied in challenging scenarios~\cite{gallego2020event, wang2025estr, zhang2024evsign, wang2025eventstr, zhou2024exact, yang2023event, yang2024event, liang2025efficient, klenk2024masked, chen2025evlight++, wang2023unsupervised, peng2020globally, zhang2021object, wan2024event, he2024microsaccade, verma2024etram, chanda2025event, gehrig2020video, rebecq2019events}. With the rapid development of multimodal large language models (MLLMs)~\cite{bai2025qwen2,zhang2024llavanextvideo,team2025minicpm4,li2023videochat, ott2024text} in recent years, event-based MLLMs~\cite{liu2025eventgpt, li2025eventvl, zhou2025llafea, chen2025let, qin2025event, zheng2023deep, li2025semantic, lu2025multimodal, zhang2025adaptive, hu2025controlevents} have increasingly demonstrated strong capabilities and generality in understanding asynchronous event streams. While video-based MLLMs have been extensively benchmarked across diverse tasks (e.g., understanding, recognition, and spatial reasoning)~\cite{fu2025video, li2024mvbench, wang2025lvbench, zhou2025mlvu, tan2025allvb, hong2025motionbench}, comprehensive evaluation benchmarks for event-based MLLMs remain largely unexplored.

Most existing event-based MLLMs achieve strong performance on their respective evaluation tasks, lacking a publicly available and unified benchmark to comprehensively assess their general capabilities across diverse dimensions. Pioneering works such as EventGPT~\cite{liu2025eventgpt} and EventVL~\cite{li2025eventvl} present instructional datasets for scene understanding and visual question answering. Subsequent efforts, such as LLaFEA~\cite{zhou2025llafea} and LET-US~\cite{chen2025let}, expand the scope to incorporate spatiotemporal localization and long-term event stream understanding. While these event-based MLLMs primarily focus on evaluating specific aspects of model capability, their assessment dimensions remain relatively limited compared with the mainstream benchmarks for video-based MLLMs (e.g., understanding, recognition, and spatial reasoning). In other words, this gap highlights the need for a unified benchmark to comprehensively evaluate event-based MLLMs across diverse dimensions. Moreover, the limited public accessibility of existing event-based benchmarks~\cite{kong2025talk2event,zhou2024exact, yang2025ezsr, zhu2018multivehicle, de2020large, perot2020learning, orchard2015converting, hu2020ddd20} further restricts fair comparison, making it challenging to accurately gauge progress in event-based MLLMs.

In this work, we introduce EventBench, a comprehensive evaluation benchmark covering 8 diverse task metrics and a large-scale event stream dataset. To achieve our objective, two main challenges require to be
addressed: (i) How can we design a systematic and comprehensive evaluation benchmark that measures capabilities across multiple dimensions, such as understanding, recognition, and spatial reasoning? How can we employ various MLLMs to validate the proposed benchmark's effectiveness across different tasks, ensuring it serves to foster progress in this topic?

To address these challenges, we introduce a comprehensive benchmark that systematically evaluates event-based MLLMs across diverse dimensions (see Fig.~\ref{fig:motivation}). It comprises 8 representative tasks covering understanding, recognition, and spatial reasoning. It integrates over 20 heterogeneous data sources drawn from both synthetic and real-world event streams to ensure rich scene diversity. Importantly, it pioneers the introduction of spatial reasoning evaluation for event-based MLLMs, filling a gap in assessing models’ 3D understanding capabilities. Furthermore, we present a fully open and publicly accessible evaluation pipeline that enables different models to be evaluated under an unified benchmark, providing the opportunity to explore this evolving research topic. In addition, we construct a large-scale and carefully curated instruction dataset (i.e., EQA-1.4M) to enhance the baseline performance of open-source event-based MLLMs. We also establish objective metrics (e.g., multiple-choice accuracy, text matching, and numerical evaluation) to ensure fairness and reproducibility.

To evaluate the effectiveness of EventBench across different tasks, we assess both state-of-the-art closed-source models (i.e., GPT-5~\cite{openai2025gpt5}, GPT-4o~\cite{openai2024gpt4o}, Gemini 2.5-Pro~\cite{comanici2025gemini}, and Gemini 2.5-Flash~\cite{comanici2025gemini}) as well as leading open-source MLLMs (i.e., Qwen2.5-VL~\cite{bai2025qwen2}, InternVL 3.5~\cite{wang2025internvl3}, MiMo-VL~\cite{team2025minicpm4}, and MiniCPM4-V~\cite{xiaomimimo2025mimo-vl}). We further benchmark open-source event-based MLLMs (i.e., EventGPT~\cite{liu2025eventgpt}). The results demonstrate that EventGPT+, trained with our large-scale instruction dataset EQA-1.4M, achieves the best overall performance. Moreover, fine-tuning open-source MLLMs on EQA-1.4M leads to substantial performance gains, highlighting the dataset’s effectiveness in enhancing event-based model performance.

Our main contributions can be summarized as follows:
\begin{itemize}[noitemsep,topsep=0pt,leftmargin=15pt]
    \item We present \emph{EventBench}, a publicly accessible evaluation benchmark with a large-scale event stream dataset, which will accelerate research on event-based MLLMs.
    \item We design \emph{a comprehensive benchmark} covering 8 diverse task metrics for understanding, recognition, and spatial reasoning, which significantly expands the range of evaluation tasks compared to existing benchmarks.
    \item We build EQA-1.4M, \emph{a large-scale dataset} of over one million synthetic and real-world event streams for event-based MLLMs, considering scene diversity across 20 data sources and various lengths of event streams.
    \item We conduct \emph{extensive experiments} to compare state-of-the-art closed-source MLLMs, open-source MLLMs, and event-based MLLMs. The experimental results provide a guiding benchmark for event-based MLLMs.
\end{itemize}

\begin{figure*}[t]
  \centering
   \includegraphics[width=\linewidth]{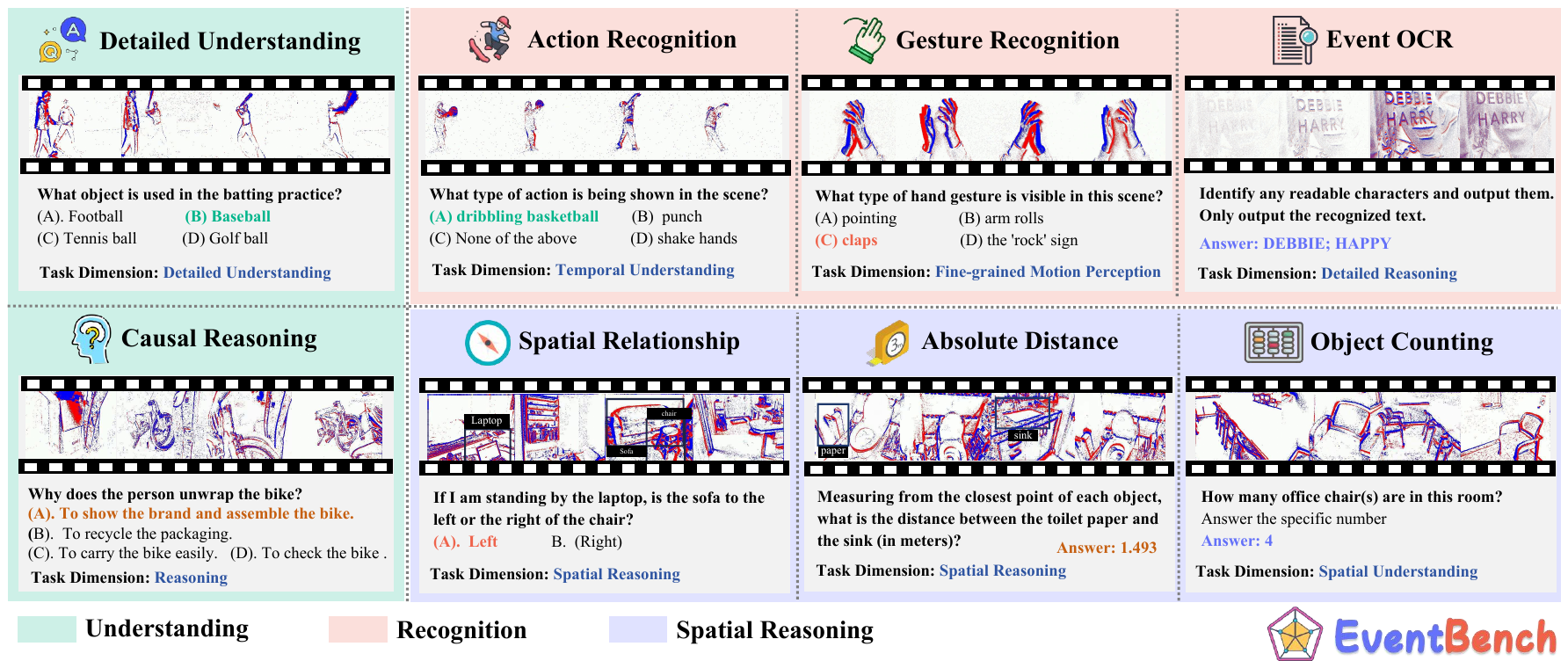}
   \caption{The comprehensive EventBench covers 8 diverse task metrics for systematically evaluating the capabilities of event-based MLLMs. These metrics can be broadly categorized into three types: understanding (i.e., detailed understanding), recognition (i.e., action recognition, gesture recognition, and event OCR), and spatial reasoning (i.e., spatial relationship, absolute distance, and object counting).}
   \label{fig:overview}
\end{figure*}

%% file: sec/2_related_works.tex
\section{Related Works}
\label{sec:related_works}

\textbf{Event-based MLLMs.} Early works (e.g., EventCLIP~\cite{wu2023eventclip} and EventBind~\cite{zhou2024eventbind}) in event-language multimodal learning tightly coupled event streams with language~\cite{kong2025talk2event, yang2025ezsr}, achieving efficient event-language alignment and enabling downstream tasks such as event retrieval and object recognition. With the rapid advancement of LLMs~\cite{zhang2023video, lin2023video, li2023videochat, maaz2023video}, event-based vision MLLMs have seen significant progress. Pioneering models, such as EventGPT~\cite{liu2025eventgpt} and EventVL~\cite{li2025eventvl}, design event-based LLMs to leverage inherent world knowledge for event stream understanding. In addition, LLaFEA~\cite{zhou2025llafea} extends region-level perception using both events and frames, while LET-US~\cite{chen2025let} explores the capabilities of event-based vision MLLMs in modeling long-duration event streams. Nevertheless, these event-based MLLMs achieve strong performance on their respective evaluation tasks, yet they lack a unified benchmark to assess their general capabilities across diverse dimensions.


\noindent \textbf{Benchmarks for Event-based MLLMs.} Effective benchmarks are vital for advancing event-based MLLMs, as they provide guidance for evaluating model capabilities and fostering new research. For instance, EventChat~\cite{liu2025eventgpt} is the first self-constructed event-based dataset for event-based MLLMs, designed to assess models’ performance in event-driven captioning and visual question answering. LLaFEA-Bench~\cite{zhou2025llafea} focuses on region-level understanding by incorporating object localization, while EVQA-Bench~\cite{chen2025let} targets long-term event stream comprehension. As shown in Table~\ref{tab:datasets}, these benchmarks provide quantitative assessments of specific abilities but remain narrow in scope, typically focusing on a limited range of tasks. Furthermore, they are often not publicly accessible, limiting reproducible comparisons across models and research on this evolving topic.


%% file: sec/3_method.tex
\begin{figure*}[t]
  \centering
   \includegraphics[width=\linewidth]{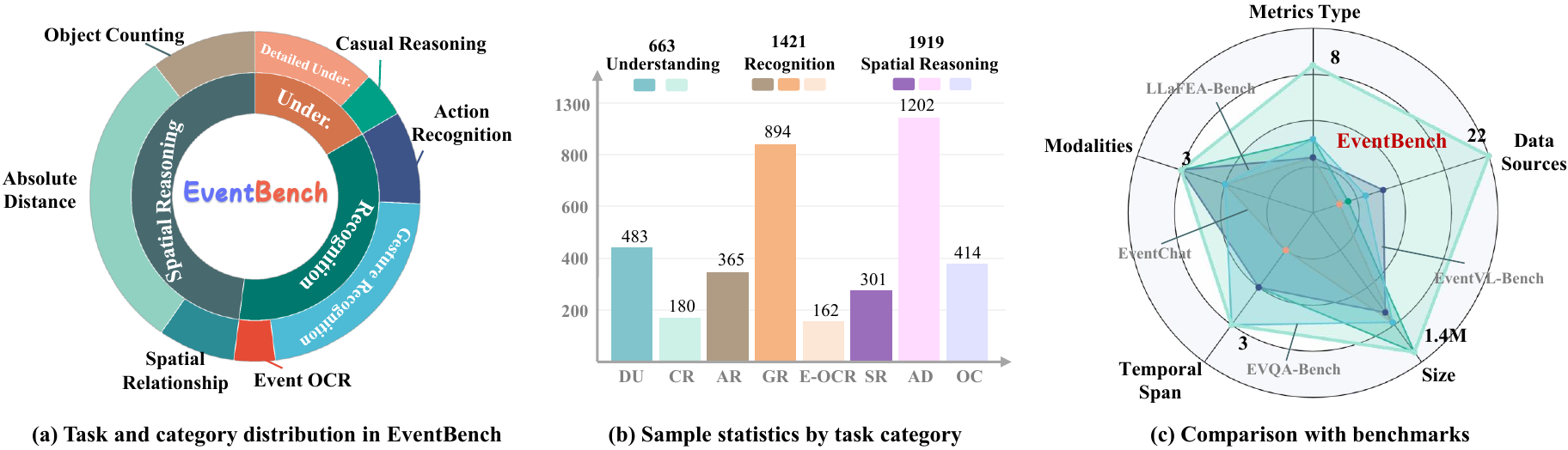}
   \caption{
Data statistics of our EventBench. (a) Task and category distribution across three groups: understanding (i.e., DU and CR), recognition (i.e., AR, GR, and E-OCR), and spatial reasoning (i.e., SR, AD, and OC). (b) Sample statistics for each task category. (c) Comparison with existing event-based benchmarks across multiple dimensions (i.e., modality, metric type, data source, size, and temporal span). Note that EventBench provides a comprehensive benchmark for systematically evaluating the capabilities of event-based MLLMs.
   }
   \label{fig:data_statistics}
\end{figure*}

\section{EventBench}
\label{sec:Method}


\subsection{Data Curation}
The data preparation process of EventBench and EQA-1.4M consists of three main stages: event stream collection and synthesis, quality assurance, and instruction generation.
Through this process, we aim to establish a comprehensive and high-quality evaluation platform specifically designed to assess the capabilities of event-based MLLMs. The detailed process is described as follows.

\noindent{\textbf{Event Stream Collection and Synthesis.}} To ensure scene diversity, we select over 20 data sources covering both real-world and synthetic event streams. These include widely used real-world event-based datasets such as DailyDVS-200 \cite{wang2024dailydvs}, HARDVS \cite{wang2024hardvs}, Bullying-10k \cite{dong2023bullying10k}, EB-HandGesture \cite{aitsam2024event}, EHWGesture \cite{amprimo2025ehwgesture}, EventSTR \cite{wang2025eventstr}, DSEC \cite{gehrig2021dsec}, and N-ImageNet \cite{kim2021n}. 
To further enhance scene diversity, we also incorporate multiple scene-centric video datasets featuring dynamic scenarios, including Kinetics-700 \cite{carreira2019short}, ActivityNet \cite{caba2015activitynet}, Charades \cite{sigurdsson2018charades}, MotionBench \cite{hong2025motionbench}, MotionSight \cite{du2025motionsight}, Wevid-10M \cite{bain2021frozen}, SportsSloMo \cite{chen2023sportsslomo}, ScanNet \cite{dai2017scannet}, PE-Data \cite{bolya2025perception}, and PLM-Data \cite{cho2025perceptionlm}. This integration ensures a balanced coverage across indoor and outdoor scenarios, providing a rich foundation for evaluation. All video datasets are subsequently converted into event streams using the V2E \cite{hu2021v2e} to build a large-scale synthesized dataset. 

\noindent\textbf{Quality Assurance.}
To ensure reliable and high-quality raw data sources, we implement a two-stage quality assurance strategy. In the first stage, we prioritize source videos exhibiting rich motion dynamics. For videos accompanied by captions, GPT-4o  is employed to verify whether the descriptions correspond to dynamic scenes, and only those meeting this criterion are retained. In the second stage, we develop an automated filtering pipeline. A subset of 10k converted samples is manually annotated for quality assessment, and these annotations are used to fine-tune a Qwen2.5-VL-7B \cite{bai2025qwen2} evaluation model. The trained model is subsequently applied to the entire dataset to automatically remove low quality samples. This process yields a high quality corpus of 112k event streams that span both task-specific and general dynamic scenarios.


\begin{table}[t]
\centering
\small
\renewcommand{\arraystretch}{1.15}
\resizebox{\columnwidth}{!}{
\begin{tabular}{c|c|c}
\toprule
\textbf{Category} & \textbf{Evaluation Dimension} & \textbf{Data Sources} \\
\midrule
\multirow{2}{*}{\textbf{Understanding}} 
& Detailed Understanding & \makecell[c]{Kinetics~\cite{carreira2019short}, SportSloMo~\cite{chen2023sportsslomo}, MotionBench~\cite{hong2025motionbench} \\ PLM-Data~\cite{cho2025perceptionlm}, PE-Data~\cite{bolya2025perception}, DSEC~\cite{gehrig2021dsec}, N-ImageNet~\cite{kim2021n}}  \\
& Causal Reasoning & ActivityNet~\cite{caba2015activitynet}, Charades~\cite{sigurdsson2018charades}, MotionSight~\cite{du2025motionsight}, WebVid-10M~\cite{bain2021frozen} \\
\midrule
\multirow{3}{*}{\textbf{Recognition}} 
& Action Recognition & DailyDVS-200~\cite{wang2024dailydvs}, Bullying-10k~\cite{dong2023bullying10k}, UCF-101~\cite{soomro2012ucf101}, HARDVS~\cite{wang2024hardvs} \\
& Gesture Recognition & EB-HandGesture~\cite{aitsam2024event}, EHWGesture~\cite{amprimo2025ehwgesture} \\
& Event-OCR & EventSTR~\cite{wang2025estr} \\
\midrule
\multirow{3}{*}{\textbf{Spatial Reasoning}} 
& Spatial Relationship & ScanNet-v2~\cite{dai2017scannet}, ARKitScenes~\cite{baruch2021arkitscenes}  \\
& Absolute Distance & ScanNet-v2, ARKitScenes  \\
& Object Counting & ScanNet-v2, ARKitScenes  \\
\bottomrule
\end{tabular}
}
\caption{Evaluation dimensions and data sources in EventBench. EventBench includes 8 representative tasks covering understanding, recognition, and spatial reasoning. It brings together over 20 data sources from both synthetic and real-world event streams.}
\label{tab:eventbench_dimensions}
\end{table}

\noindent{\textbf{Instruction Generation. }}We construct a unified instruction generation pipeline that formulates diverse tasks under a consistent QA paradigm. The 8 tasks in EventBench are organized into three categories: understanding, recognition, and spatial reasoning, each reflecting different aspects of event-based MLLM capabilities.

For understanding tasks (i.e., detailed understanding and causal reasoning), GPT-5 is prompted to synthesize multiple choice question–answer (MCQA) pairs based on the RGB videos paired with event streams. A reasoning-oriented prompting strategy guides the model to generate explicit temporal and causal inference chains, while human verification ensures factual accuracy and modality alignment. For recognition tasks (i.e., action recognition, gesture recognition, and event OCR), structured annotations are reformulated into natural language QA form. Each instruction concisely defines the recognition objective and contextual scope, and the candidate labels are converted into multiple choice options, thereby unifying recognition and understanding within a shared evaluation format.

Beyond understanding and recognition, we further introduce spatial reasoning tasks to evaluate models’ 3D geometric reasoning. We design three tasks (i.e., object counting, absolute distance estimation, and spatial relational reasoning), each grounded in the intrinsic geometry of the scene and constructed from instance-level annotations in ScanNet-V2~\cite{dai2017scannet} and ARKitScene~\cite{baruch2021arkitscenes}. The dataset provides metrically accurate spatial layouts that ensure geometric consistency, enabling physically interpretable and quantitatively verifiable evaluation of spatial reasoning ability. To ensure both reliability and consistency of the constructed data, we adhere to three guiding principles:  
(i) \emph{Geometry consistent}, where all spatial relationships are derived directly from metric 3D coordinates rather than image projections, ensuring geometric fidelity; (ii) \emph{Instance level}, where each query is anchored to uniquely identified object instances, avoiding semantic ambiguity across categories; (iii) \emph{Human centric}, where spatial relations are represented in egocentric form, aligning geometric definitions with intuitive human perception.

\subsection{Dataset Statistics}
EventBench encompasses 8 diverse task metrics for event-based MLLMs, comprising 112k event streams and 4,003 QA pairs. Each sample includes an event stream, its paired video, and a carefully designed textual instruction. In contrast to previous event-based benchmarks \cite{chen2025let, li2025eventvl,liu2025eventgpt,zhou2025llafea} that define only limited evaluation scopes, EventBench establishes a broader and more systematic benchmark unifying understanding, recognition, and spatial reasoning. The overall task composition is illustrated in Fig.~\ref{fig:data_statistics}, which summarizes the hierarchical distribution across categories and individual task scales. We further analyze the attributes of the benchmark from the following three perspectives.

\emph{Category Diversity.} The benchmark is hierarchically organized into three major categories that reflect progressive levels of event stream understanding. The understanding category contains 663 QA pairs, which emphasize temporal comprehension and causal inference. The recognition category comprises 1,421 QA pairs focusing on the model’s adaptability to real-world event-based vision tasks. The spatial reasoning category consists of 1,919 QA pairs designed to evaluate 3D geometric reasoning.

\begin{figure}[t]
  \centering
   \includegraphics[width=\linewidth]{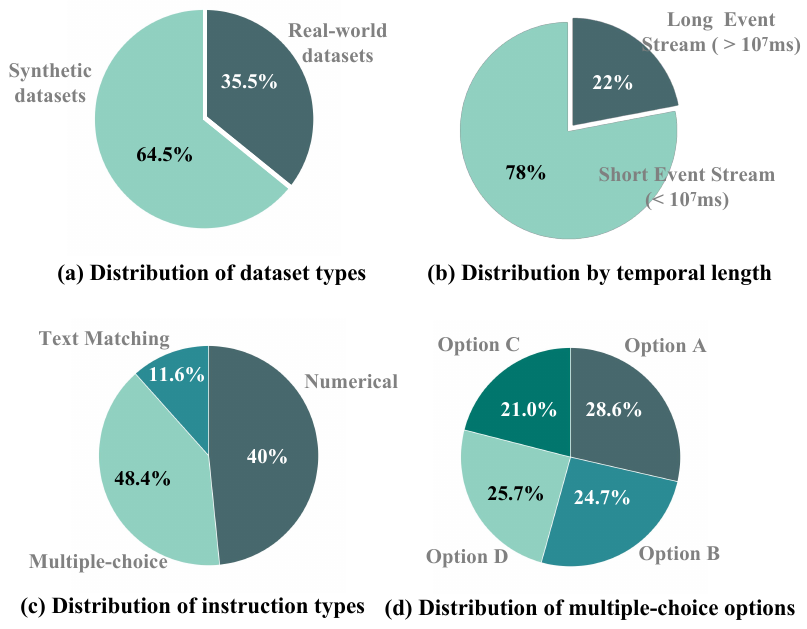}
   \caption{Data statistics in our EventBench across multiple dimensions. (a) Distribution of synthetic and real-world datasets. (b) Proportions of event streams by temporal length. (c)-(d) Distributions of instruction types and multiple-choice options.}
   \label{fig:pie_graph}
\end{figure}

\emph{Temporal Duration.} To capture both short-term and long-term motion dynamics, EventBench integrates event streams with diverse temporal lengths divided by a threshold of 10 second. As shown in Fig.~\ref{fig:pie_graph}, long event streams account for \(22\)\% of all samples, while short ones make up the remaining \(78\)\%. This temporal distribution provides balanced and comprehensive coverage across motion scales, thereby enabling consistent evaluation of model robustness under both short-term and long-term event streams.

\emph{Instruction Composition.} The benchmark employs three types of objective question formats, including multiple choice, numerical, and text match. Each type contains a single correct answer to ensure reproducibility and precision in evaluation. Multiple-choice questions constitute \(48.4\)\% of the dataset, numerical questions \(40.0\)\%, and text-match questions \(11.6\)\%. As depicted in Fig.~\ref{fig:pie_graph}, the multiple-choice options are evenly distributed across A, B, C, and D with proportions of \(28.6\)\%, \(24.7\)\%, \(21.0\)\%, and \(25.7\)\%, respectively. This balanced configuration minimizes bias and ensures that scores reflect genuine reasoning ability.

\begin{table*}[t]
\centering
\renewcommand{\arraystretch}{1.15}
\setlength{\tabcolsep}{2pt}
\caption{Evaluation results on EventBench, showing three categories of models: closed-source MLLMs, open-source MLLMs (SFT), and open-source event-based MLLMs. The 8 metrics correspond to: detailed understanding (DU), causal reasoning (CR), action recognition (AR), gesture recognition (GR), event-ocr (E-OCR), spatial relationship (SR), absolute distance (AD), and object counting (OC).}
\resizebox{\textwidth}{!}{
\begin{tabular}{l>{\centering\arraybackslash}p{0.10\textwidth}|>{\centering\arraybackslash}p{0.15\textwidth}|*{2}{>{\centering\arraybackslash}p{0.07\textwidth}}|*{3}{>{\centering\arraybackslash}p{0.075\textwidth}}|*{3}{>{\centering\arraybackslash}p{0.075\textwidth}}|>{\centering\arraybackslash}p{0.085\textwidth}}
\toprule
\multirow{2}{*}{\textbf{Model}} &
\multirow{2}{*}{\textbf{Params}} &
\multirow{2}{*}{\textbf{Backbone}} &
\multicolumn{2}{c}{\textbf{Understanding}} &
\multicolumn{3}{c}{\textbf{Recognition}} &
\multicolumn{3}{c}{\textbf{Spatial Reasoning}} &
\multirow{2}{*}{\textbf{Overall}} \\
\cmidrule(lr){4-5}\cmidrule(lr){6-8}\cmidrule(l){9-11}
 &  &  & DU & CR & AR & GR & E-OCR & SR & AD & OC &  \\
\midrule
\rowcolor[HTML]{E9EAFE}
\multicolumn{12}{l}{\textbf{\textcolor{blue}{\textbullet~Close-Source MLLMs}}} \\
GPT-5\cite{openai2025gpt5} & - & 2025-08-07 &\(68.5\)  &\(\mathbf{70.1}\)  &\(46.0\)  &\(49.8\)  &\(22.2\)  &\(\mathbf{47.8}\)  &\(30.7\)  &\(\mathbf{25.6}\)  &\(\mathbf{45.1}\)  \\
GPT-4o\cite{openai2024gpt4o} & - & 2024-11-20 &\(62.5\)  &\(65.8\)  &\(\mathbf{55.3}\)  &\(48.6\)  &\(19.1\)  &\(45.5\)  &\(17.8\)  &\(21.5\)  &\(42.0\)  \\
Gemini2.5-pro\cite{comanici2025gemini}& - & 2025-06-17
 &\(61.3\)  &\(63.9\)  &\(44.1\)  &\(\mathbf{49.9}\)  &\(19.1\)  &\(40.5\)  &\(17.5\)  &\(16.9\)  &\(39.2\)  \\
Gemini2.5-flash\cite{comanici2025gemini} & - & 2025-06-17
 &\(54.7\)  &\(61.1\)  &\(42.7\)  &\(43.4\)  &\(21.0\)  &\(25.6\)  &\(15.6\)  &\(14.5\)  &\(34.8\)  \\
Doubao-Seed-1.6-vision & - & - &\(\mathbf{70.3}\)  &\(65.2\)  &\(46.6\)  &\(45.4\)  &\(\mathbf{45.1}\)  &\(11.6\)  &\(\mathbf{36.3}\)  &\(23.0\)  &\(42.9\)  \\
\midrule
\rowcolor[HTML]{DFF6F0}
\multicolumn{12}{l}{\textbf{\textcolor{green!50!black}{\textbullet~Open-Source MLLMs (SFT)}}} \\
Qwen2.5-VL\cite{bai2025qwen2} & 3B & Qwen2.5
  &\(59.8\)  &\(65.6\)  &\(52.9\)  &\(35.2\)  &\(\mathbf{45.6}\)  &\(\mathbf{44.9}\)  &\(31.9\)  &\(59.9\)  &\(49.5\)  \\
InternVL3.5\cite{wang2025internvl3} & 4B & Qwen3
  &\(73.1\)  &\(\mathbf{75.0}\)  &\(\mathbf{69.3}\)  &\(\mathbf{45.2}\)  &\(38.9\)  &\(40.5\)  &\(\mathbf{46.8}\)  &\(\mathbf{61.1}\)  &\(\mathbf{56.2}\)  \\
MiniCPM-V-4\cite{team2025minicpm4} & 4.1B & MiniCPM4
  &\(\mathbf{72.9}\)  &\(63.9\)  &\(58.9\)  &\(37.1\)  &\(44.4\)  &\(44.2\)  &\(43.1\)  &\(55.6\)  &\(52.5\)  \\
MiMo-VL\cite{xiaomimimo2025mimo-vl} & 7B & MiMo
  &\(71.8\)  &\(68.3\)  &\(57.3\)  &\(38.6\)  &\(31.5\)  &\(44.5\)  &\(45.2\)  &\(61.0\)  &\(52.3\)  \\
\midrule
\rowcolor[HTML]{FEE8E4}
\multicolumn{12}{l}{\textbf{\textcolor{purple!80!black}{\textbullet~Open-Source Event-Based MLLMs}}} \\
EventGPT\cite{liu2025eventgpt} & 7B & Vicuna-v1.5  &\(76.8\)  &\(79.2\)  &\(78.6\)  &\(57.2\)  &\(52.4\)  &\(49.8\)  &\(52.6\)  &\(61.5\)  &\(63.5\)  \\
\textbf{EventGPT+(B) (Ours*)} & 2B & Qwen2.5  &\(78.3\)  &\(78.2\)  &\(79.8\)  &\(55.5\)  &\(53.6\)  &\(50.2\)  &\(51.9\)  &\(62.8\)  &\(63.8\)  \\

\textbf{EventGPT+(L) (Ours*)} & 7B & Qwen2.5  &\(\mathbf{79.1}\)  &\(\mathbf{79.6}\)  &\(\mathbf{81.6}\)  &\(\mathbf{58.7}\)  &\(\mathbf{55.4}\)  &\(\mathbf{53.2}\)  &\(\mathbf{52.2}\)  &\(\mathbf{64.6}\)  &\(\mathbf{65.5}\)  \\
\bottomrule
\label{tab:main_results}
\end{tabular}
}
\end{table*}

\subsection{Task Definition}

To establish a unified benchmark for systematically evaluating event-based vision MLLMs, we define a structured set of eight tasks covering understanding, recognition, and spatial reasoning, as detailed below.


\noindent\textbf{Detailed Understanding (DU).}  
This task evaluates a model’s ability to interpret scene content in event streams, requiring precise identification of objects, activities, and contextual cues. Typical questions include identifying what is happening in the scene or recognizing key elements (\emph{``What is the person doing?''}). This capability reflects the model’s fundamental competence in understanding fine-grained semantics within dynamic event streams.

\noindent\textbf{Causal Reasoning (CR).}
Inferring causal relationships from event streams requires interpreting temporal dependencies and intention cues (\emph{``Why did the object fall?’’}). This task evaluates a model’s ability to uncover the underlying mechanisms behind actions and outcomes.

\noindent\textbf{Action Recognition (AR).}
Identifying human actions from event stream requires detecting motion dynamics and inferring activity categories (\emph{``What action is being performed?''}). This task evaluates a model’s ability to capture coherent temporal dynamics under rapid motion.


\noindent{\textbf{Gesture Recognition (GR).}} Identifying human actions from the event stream requires detecting motion dynamics and inferring activity categories (\emph{``What gesture is being performed?’’}). This task evaluates a model’s ability to capture temporal continuity under rapid motion, which is essential for reliable event-driven activity understanding.

\noindent\textbf{Event OCR (E-OCR).}
Recognizing textual content in event streams requires identifying characters and reading dynamic text under highly challenging visual conditions (\emph{``What text appears in the scene?’’}). Since event cameras possess high temporal resolution and high dynamic range, they offer a distinct advantage for overcoming specific challenges like fast motion blur and sudden illumination changes. This enables robust text recognition in scenarios where conventional cameras fail.


\noindent\textbf{Spatial Relationship (SR).}
Understanding relative object positions within 3D space requires determining where one object lies with respect to another (\emph{``Where is X relative to Y?’’}). This capability depends on grounding spatial cues and geometric structure, and is essential for robust 3D spatial reasoning in dynamic event stream environments.


\noindent\textbf{Absolute Distance (AD).}  
This task measures a model’s ability to estimate metric distances between scene entities using spatial cues inferred from the event stream (\emph{``How far apart are A and B?’’}). It evaluates quantitative spatial reasoning and geometric awareness, reflecting the model’s ability to internalize depth cues from event representations.

\noindent\textbf{Object Counting (OC).}
Estimating the number of objects in an event stream (\emph{``How many instances are there?’’}) requires managing cluttered scenes and leveraging fine-grained spatial cues. The task evaluates numerical reasoning coupled with event-based spatial perception.

\subsection{Comparison with Other Benchmarks}
To highlight the advancements of the new benchmark, we compare it with related event-based datasets and benchmarks in Table~\ref{tab:datasets}. It indicates that our EventBench is the first open-source and comprehensive benchmark for event-based MLLMs. More precisely, EventBench incorporates 8 distinct metric types, effectively doubling the categorical coverage of the most competitive alternatives. Moreover, while existing benchmarks are often limited to short event sequences, EventBench offers a versatile suite of data with short, medium, and long-term event streams, facilitating a more thorough investigation of temporal modeling abilities.

Overall, the combination of a professionally designed instruction set and a large-scale event-based dataset makes EventBench a competitive benchmark with several key features: (i) Openness: all raw event streams and task instructions across eight evaluation metrics are publicly released; (ii) Task diversity: it covers a wide range of understanding, recognition, and spatial reasoning tasks, enabling comprehensive evaluation of model capabilities; (iii) Spatial integration: it pioneers the inclusion of 3D spatial reasoning tasks for event-based MLLMs; and (iv) Data scale: it includes a training set of over one million event–text pairs, supporting large-scale training and evaluation.

%% file: sec/4_experiment.tex

\begin{figure}[t]
  \centering
   \includegraphics[width=\linewidth]{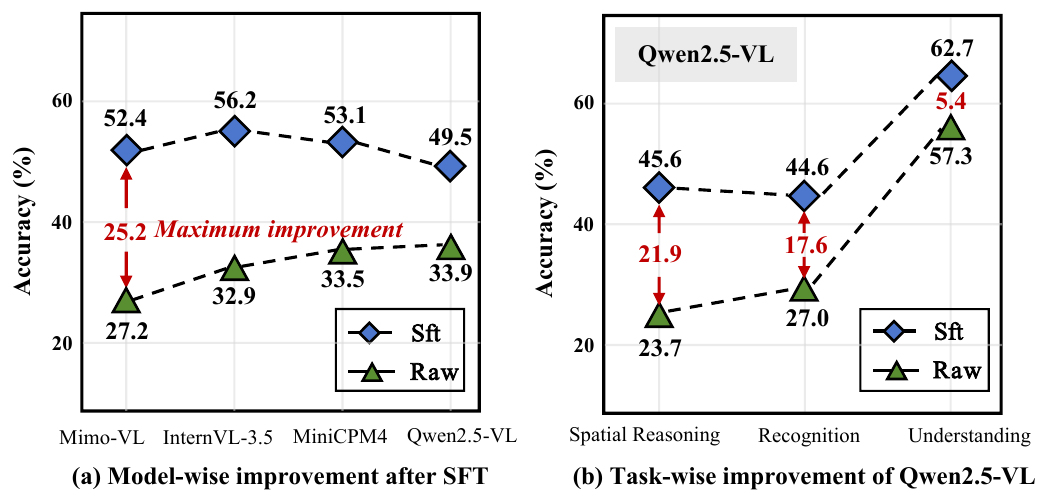}
   \caption{
Overall performance improvement on EventBench after SFT training.  (a) Model-wise improvement with a 
\(25.2\)\% maximum gain. (b) Task-wise improvement of Qwen2.5-VL across spatial reasoning, downstream, and open-domain QA tasks.
   }
   \label{fig:performance_improvement}
\end{figure}

\section{Experiments}
\subsection{Implementation Details}
To evaluate our EventBench, we select four closed-source models (i.e., GPT-5, GPT-4o, Gemini-2.5-Pro, Gemini-2.5-Flash, and Doubao-Seed-1.6-Vision) and several state-of-the-art open-source models (i.e., Qwen2.5-VL, InternVL-3.5, MiniCPM4, and MiMo-VL), and the only publicly event-based MLLM (i.e., EventGPT) and our baseline model EventGPT+. For closed-source models, evaluation is conducted on EventBench by representing event streams as event videos sampled at 1 FPS. To ensure a fair comparison, all open-source models are fine-tuned on our EQA-1.4M dataset using the same event video inputs at 1 FPS, while event-based models are fine-tuned on raw event streams. All experiments are performed on 8 NVIDIA A800 GPUs, with API-based evaluation for closed-source models and single-GPU inference for all open-source settings.

\begin{figure}[t]
  \centering
   \includegraphics[width=\linewidth]{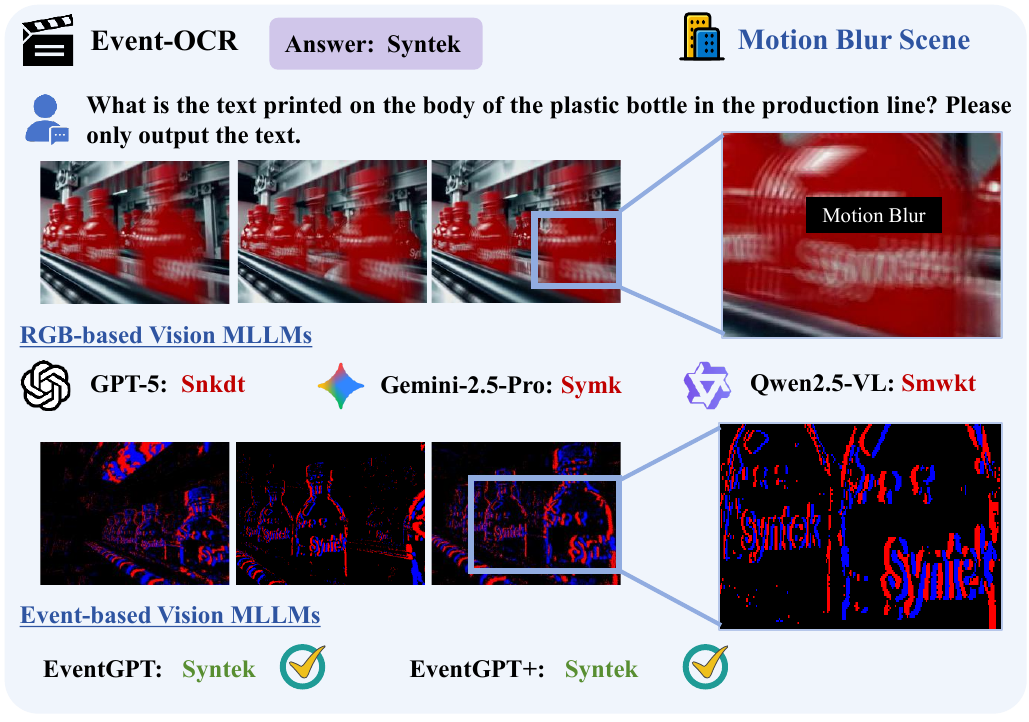}
   \caption{
Comparison of frame-based and event-based MLLMs on the Event-OCR task under motion blur scenarios. Event-based models (i.e., EventGPT and EventGPT+) successfully recognize “Syntek”, while RGB-based models fail due to motion blur.
   }
   \label{fig:motion_blur}
\end{figure}

\begin{table}[t]
\centering
\caption{Performance of open-source MLLMs before and after SFT on EQA-1.4M, evaluated across on EventBench.}
\small
\renewcommand{\arraystretch}{1.3}
\setlength{\tabcolsep}{5pt} 
\resizebox{\linewidth}{!}{
\begin{tabular}{l|cc|cc}
\toprule
\textbf{Task} & \textbf{Qwen-VL} & \textbf{SFT} & \textbf{Intern-VL} & \textbf{SFT} \\
\midrule
\rowcolor{gray!12}
\multicolumn{5}{l}{\textbf{Understanding}} \\
\textbf{Detailed Understanding} &\(53.8\) & \(\mathbf{59.8}\)\textcolor[RGB]{136,0,204}{\small{(\(+6.0\))}} &\(47.2\) & \(\mathbf{73.1}\)\textcolor[RGB]{136,0,204}{\small{(\(+25.9\))}} \\
\textbf{Causal Reasoning} &\(59.8\) & \(\mathbf{65.6}\)\textcolor[RGB]{136,0,204}{\small{(\(+5.8\))}} &\(63.2\) & \(\mathbf{75.0}\)\textcolor[RGB]{136,0,204}{\small{(\(+11.8\))}} \\
\midrule
\rowcolor{gray!12}
\multicolumn{5}{l}{\textbf{Recognition}} \\
\textbf{Action Recognition} &\(29.3\) & \(\mathbf{52.9}\)\textcolor[RGB]{136,0,204}{\small{(\(+23.6\))}} &\(24.1\) & \(\mathbf{69.3}\)\textcolor[RGB]{136,0,204}{\small{(\(+45.2\))}} \\
\textbf{Gesture Recognition} &\(31.5\) & \(\mathbf{35.2}\)\textcolor[RGB]{136,0,204}{\small{(\(+3.7\))}} &\(32.9\) & \(\mathbf{45.2}\)\textcolor[RGB]{136,0,204}{\small{(\(+12.3\))}} \\
\textbf{Event OCR} &\(20.2\) & \(\mathbf{45.6}\)\textcolor[RGB]{136,0,204}{\small{(\(+25.4\))}} &\(22.8\) & \(\mathbf{38.9}\)\textcolor[RGB]{136,0,204}{\small{(\(+16.1\))}} \\
\midrule
\rowcolor{gray!12}
\multicolumn{5}{l}{\textbf{Spatial Reasoning}} \\
\textbf{Spatial Relationship} &\(38.8\) & \(\mathbf{44.9}\)\textcolor[RGB]{136,0,204}{\small{(\(+6.1\))}} &\(36.8\) & \(\mathbf{40.5}\)\textcolor[RGB]{136,0,204}{\small{(\(+3.7\))}} \\
\textbf{Absolute Distance} &\(21.7\) & \(\mathbf{31.9}\)\textcolor[RGB]{136,0,204}{\small{(\(+10.2\))}} &\(18.9\) & \(\mathbf{46.8}\)\textcolor[RGB]{136,0,204}{\small{(\(+27.9\))}} \\
\textbf{Object Counting} &\(10.8\) & \(\mathbf{59.9}\)\textcolor[RGB]{136,0,204}{\small{(\(+49.1\))}} &\(16.9\) & \(\mathbf{61.1}\)\textcolor[RGB]{136,0,204}{\small{(\(+44.2\))}} \\
\bottomrule
\end{tabular}
}
\vspace{-4pt}
\label{tab:vl_sft_comparison}
\end{table}

\subsection{Evaluation On EventBench}
\begin{figure}[t]
  \centering
   \includegraphics[width=\linewidth]{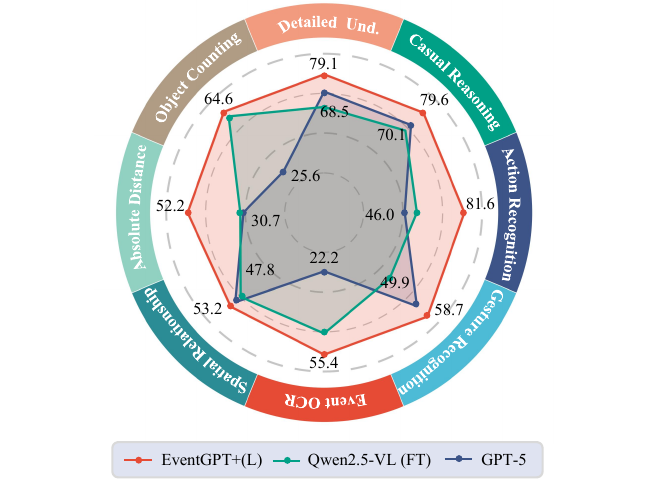}
   \caption{
Comparison of EventGPT+(L), Qwen2.5-VL (FT), and GPT-5 on 8 evaluation tasks in EventBench. EventGPT+(L) exhibits the best overall performance in EventBench.
   }
   \label{fig:lidar_graph}
\end{figure}

\noindent{\textbf{Quantitative Results.}} We conduct an extensive evaluation on EventBench to assess the generalization and reasoning capabilities of both video-based and event-based MLLMs. As summarized in Table~\ref{tab:main_results}, the evaluated models are grouped into three categories: advanced closed-source MLLMs, open-source video-based MLLMs fine-tuned on EQA-1.4M, and the only publicly available event-based MLLM (i.e., EventGPT). Meanwhile, our baseline EventGPT+ is designed to handle long event streams via dynamic event bin selection. In addition, Fig.~\ref{fig:lidar_graph} presents a radar-style comparison of model performance across EventBench, highlighting the distinctions among commercial models, open-source MLLMs, and event-based vision MLLMs. Specifically, GPT-5, Qwen2.5-VL, and EventGPT+ serve as representatives of these three categories, demonstrating the superior performance of EventGPT+(L) across multiple event understanding dimensions.

Among advanced closed-source models, GPT-5 achieves the highest overall accuracy, outperforming Gemini-2.5-Flash by \(10.3\)\%. While their strong general visual-language reasoning abilities, all closed-source models perform notably worse than the open-source video MLLMs after supervised fine-tuning on EQA-1.4M. This highlights the domain gap between standard video understanding and event stream perception. Within the open-source video-based category, InternVL-3.5 demonstrates the best performance, achieving an accuracy of \(56.2\)\%. Other models, such as Qwen2.5-VL and MiniCPM4, show consistent performance across tasks, particularly on general reasoning and action recognition. Fine-tuning on EQA-1.4M improves temporal reasoning and motion perception, indicating its effectiveness in bridging cross-modal gaps.

Event-based MLLMs exhibit a significant advantage across nearly all categories, demonstrating clear superiority in spatiotemporal reasoning. Our EventGPT+ achieves the best overall performance among all models, surpassing GPT-5 by \(20.4\)\% and outperforming the best fine-tuned open-source video MLLM by \(9.3\)\%. This indicates that directly modeling raw event streams, allows the model to better capture high temporal resolution motion cues critical for event understanding. To further validate this advantage, we compare EventGPT+(L) with the only existing open-source event-based MLLMs (i.e., EventGPT). For a fair comparison, EventGPT is further fine-tuned on EQA-1.4M, with its fixed event bin constraint removed to support long duration event streams. Under identical conditions, EventGPT+(L) consistently outperforms EventGPT across all categories, achieving an overall improvement of \(2.0\)\%. 

\begin{figure}[t]
  \centering
   \includegraphics[width=\linewidth]{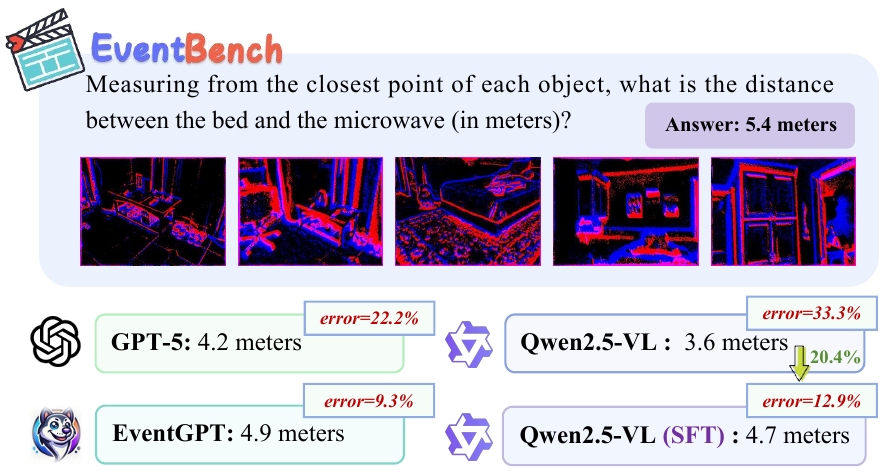}
   \caption{
Comparison of model performance on the Absolute Distance task in EventBench. EventGPT reaches a \(9.3\%\) error, and Qwen2.5-VL reduces its error by \(20.4\%\) after SFT.
   }
   \label{fig:abs_distance}
\end{figure}

\noindent\textbf{Qualitative Analysis.} 
We further conduct qualitative analyses on challenging scenarios from EventBench. 
As illustrated in Fig.~\ref{fig:motion_blur}, under severe motion blur, RGB-based models suffer from significant image degradation, which induces noticeable hallucinations across both advanced proprietary models such as GPT-5 and Gemini-2.5-Pro, and strong open-source models like Qwen2.5-VL. 
In contrast, event-based MLLMs exhibit higher robustness, effectively preserving fine-grained motion structures and temporal consistency in high-speed environments. 
Moreover, Fig.~\ref{fig:abs_distance} presents a comparison of the Absolute Distance task in EventBench. 
While most models reveal inherent limitations in spatial reasoning, EventGPT+ achieves the most accurate distance estimation with an error of \(9.3\%\). 
Qwen2.5-VL also demonstrates a clear improvement after supervised fine-tuning, reducing its error from \(33.3\%\) to \(12.9\%\).

\noindent{\textbf{Effect of Fine-tuning on EventBench.}} To ensure a fair comparison, some representative open-source video MLLMs (i.e., Qwen2.5-VL and InternVL-3.5) are fine-tuned on the EQA-1.4M dataset using event videos as input. Table~\ref{tab:vl_sft_comparison} summarizes the performance of Qwen2.5-VL and InternVL-3.5 on EventBench before and after fine-tuning. Fine-tuning yields consistent gains across the three evaluation categories (i.e., understanding, recognition, and spatial reasoning). In particular, Qwen2.5-VL attains improvements of \(49.1\)\%, \(10.2\)\%, and \(6.1\)\% for three spatial reasoning tasks (i.e., object counting, absolute distance, and spatial relationship), respectively, meanwhile InternVL-3.5 achieves gains of \(44.2\)\%, \(27.9\)\%, and \(3.7\)\% on the same tasks. Although fine-tuning markedly enhances the temporal and spatial reasoning abilities of video-based MLLMs, models that directly operate on raw event streams still deliver superior overall performance on EventBench.

\noindent{\textbf{Effect of Sampling Interval.}} To assess the influence of the sampling interval on the final performance, we evaluate four settings with $\Delta t$ $\in$ $\{0.5, 1.0, 1.5, 2.0\}$. As shown in Table~\ref{tab:sampling_interval}, varying $\Delta t$ yields only minor fluctuations (within $2.5\%$) across all models, indicating strong robustness to temporal sampling and confirming the stability of EventBench. Given this consistent behavior and considering the balance between accuracy and computational efficiency, we recommend adopting $\Delta t$ $=$ $ 1.0$ as the default configuration for evaluating event-based MLLMs using EventBench.

\noindent\textbf{Effect of Aggregation Window Size.} 
To further examine the impact of temporal granularity, we analyze the effect of the aggregation window size on EventBench. 
As shown in Table~\ref{tab:aggregation_windows_size}, we evaluate four configurations with window sizes of \(10\,ms\), \(20\,ms\), \(30\,ms\), and \(40\,ms\). 
The results show that a \(20\, ms\) window achieves the best overall performance. 
When the window size is too small, the number of events within each bin becomes insufficient, resulting in underpopulated or even empty windows that weaken temporal feature integration. 
In contrast, overly large windows produce coarser temporal segmentation, blurring fine-grained spatiotemporal cues and ultimately degrading the performance on tasks sensitive to temporal dynamics.

\begin{table}[t]
\centering
\caption{Performance comparison of open-source and event-based multimodal large language models (MLLMs) under different sampling intervals $\Delta t$ to assess the impact of temporal granularity.}
\small
\renewcommand{\arraystretch}{1.2}
\resizebox{\linewidth}{!}{
\begin{tabular}{l@{\hspace{3pt}}|c|@{\hspace{3pt}}c@{\hspace{3pt}}|cccc}
\toprule
\multirow{2}{*}{\textbf{Model}} & 
\multirow{2}{*}{\textbf{Venue}} & 
\multirow{2}{*}{\textbf{Params}} & 
\multicolumn{4}{c}{\textbf{Sampling Interval ($\Delta t$)}} \\
\cmidrule(lr){4-7}
 &  &  & $\Delta t =0.5$ & $\Delta t =1.0$ & $\Delta t =1.5$ & $\Delta t =2.0$ \\
\midrule
\rowcolor{gray!12}
\multicolumn{7}{c}{\textbf{Open-sources MLLMs}} \\
\textbf{Qwen2.5-VL}   & arXiv'25 & 3B &\(\mathbf{49.7}\)  &\(49.5\)  &\(48.9\)  &\(48.7\)  \\
\textbf{InternVL-3.5} & arXiv'25  & 4B &\(\mathbf{56.5}\)  &\(56.2\)  &\(55.9\)  &\(55.6\)  \\
\midrule
\rowcolor{gray!12}
\multicolumn{7}{c}{\textbf{Event-based vision MLLMs}} \\
\textbf{EventGPT}     & CVPR'25  & 7B &\(62.6\)  &\(\mathbf{63.5}\)  &\(62.4\)  &\(62.2\)  \\
\textbf{EventGPT+(L)}  & Ours     & 7B &\(64.9\)  &\(\mathbf{65.5}\)  &\(64.3\)  &\(63.9\)  \\
\bottomrule
\end{tabular}
}
\label{tab:sampling_interval}
\end{table}

\begin{table}[t]
\centering
\caption{Performance comparison of open-source and event-based MLLMs under different temporal aggregation window sizes to assess the effect of event accumulation granularity.}
\small
\renewcommand{\arraystretch}{1.2}
\resizebox{\linewidth}{!}{
\begin{tabular}{l@{\hspace{3pt}}|c|@{\hspace{3pt}}c@{\hspace{3pt}}|cccc}
\toprule
\multirow{2}{*}{\textbf{Model}} & 
\multirow{2}{*}{\textbf{Venue}} & 
\multirow{2}{*}{\textbf{Params}} & 
\multicolumn{4}{c}{\textbf{Aggregation Window Size}} \\
\cmidrule(lr){4-7}
 &  &  & 10\,ms & 20\,ms & 30\,ms & 40\,ms \\
\midrule
\rowcolor{gray!12}
\multicolumn{7}{c}{\textbf{Open-source MLLMs}} \\
\textbf{Qwen2.5-VL}   & arXiv'25 & 3B &\(49.2\)  & \(\mathbf{49.5}\) &\(48.6\)  &\(48.2\)  \\
\textbf{InternVL-3.5} & arXiv'25 & 4B &\(55.6\)  & \(\mathbf{56.2}\) &\(54.9\)  &\(54.3\)  \\
\midrule
\rowcolor{gray!12}
\multicolumn{7}{c}{\textbf{Event-based Vision MLLMs}} \\
\textbf{EventGPT}     & CVPR'25  & 7B &\(62.7\)  &\(\mathbf{63.5}\)  &\(62.3\)  &\(62.1\)  \\
\textbf{EventGPT+(L)}  & Ours     & 7B &\(64.8\)  & \(\mathbf{65.5}\) &\(64.5\)  &\(63.9\)  \\
\bottomrule
\end{tabular}
}
\label{tab:aggregation_windows_size}
\end{table}

%% file: sec/5_conclusion.tex
\section{Conclusion}

In this paper, we introduced EventBench, a fully open and comprehensive benchmark for systematically evaluating the capabilities of event-based MLLMs. EventBench stands as a competitive benchmark compared with existing event-based benchmarks, offering several key advantages: openness, broad task diversity, integration of spatial reasoning tasks, and a large data scale. Extensive evaluations were conducted on state-of-the-art proprietary models, open-source RGB-based MLLMs, and event-based MLLMs. The results reveal that current event-based MLLMs perform well in event stream understanding, yet still struggle with fine-grained recognition and spatial reasoning. We believe that EventBench will help drive further progress in event-based MLLMs by providing a unified benchmark for fair comparison and consistent evaluation.


%% file: sec/X_suppl.tex
\clearpage
\setcounter{page}{1}
\setcounter{section}{0}
\maketitlesupplementary

\section{Event Camera Sensing Mechanism}
\label{sec:Sensing Mechanism}
Event cameras, often referred to as dynamic vision sensors (DVS), operate by emitting asynchronous events in response to changes in pixel-wise log-intensity rather than capturing absolute brightness values. This sensing principle, which is reminiscent of the signal transduction mechanisms of biological retinas, stands in sharp contrast to conventional frame-based imaging, where measurements are acquired at fixed temporal intervals. An event $e_n$ is generated at pixel $\mathbf{u}_n = (x_n, y_n)$ once the variation in logarithmic luminance surpasses a contrast threshold $C$:
\begin{equation}
\log\!\left(\frac{I(x_n, y_n, t_n)}{I(x_n, y_n, t_n-\Delta t)}\right) = p_n C,
\end{equation}
where $p_n \in \{+1,-1\}$ denotes the polarity of the brightness change and $\Delta t$ indicates the inter-event time at the same location. The resulting observation sequence can thus be written as
\begin{equation}
\mathcal{E} = \{(x_n, y_n, t_n, p_n)\}_{n=1}^{N}.
\end{equation}

Due to their sparse output and microsecond-level temporal resolution, DVS sensors achieve extremely low latency and maintain reliable performance under rapid motion as well as challenging illumination conditions (e.g., low-light environments). These advantages are further enhanced by their inherently high dynamic range, which typically exceeds 120 dB. Despite these favorable sensing characteristics, existing evaluation resources provide only partial coverage of the capabilities required to assess event-based multimodal large language models (MLLMs). To address this gap, EventBench introduces a unified and systematically constructed benchmark that evaluates event-based MLLMs across multiple dimensions (i.e., understanding, recognition, and spatial reasoning) in diverse real-world scenarios.


\section{Experimental Analysis}

We conduct further analysis to reveal the key factors influencing the performance of event-based MLLMs: \emph{(i) model strengths and limitations}, \emph{(ii) challenges in spatial reasoning}, and \emph{(iii) the impact of SFT on the modality gap.}.

\noindent\textbf{How Do Event-based MLLMs Perform Across Tasks?} To examine the overall performance of event-based MLLMs across diverse tasks, we evaluate a wide range of models as reported in Tab.~\ref{tab:main_results}. Our analysis reveals two major observations. (1) \emph{Event-based MLLMs perform well on general understanding tasks.} As shown in Tab.~\ref{tab:main_results}, the state-of-the-art event-based model (i.e., EventGPT+(L)) achieves an average accuracy of \(79.4\)\% on understanding tasks, substantially outperforming strong closed-source and open-source MLLMs. This indicates that event-based MLLMs exhibit strong capability in generic scene understanding. (2) \emph{Event-based MLLMs underperform on domain-specific tasks.} Despite their strong general abilities, event-based MLLMs still struggle with domain-specific tasks (e.g., gesture recognition, event-OCR, and spatial reasoning). This suggests that the world knowledge embedded in LLMs facilitates general event understanding and enables certain zero-shot capabilities, yet remains insufficient for specialized recognition and spatial reasoning tasks.

\begin{table*}[t]
\centering
\renewcommand{\arraystretch}{1.15}
\setlength{\tabcolsep}{2pt}
\caption{Evaluation results on EventBench for additional closed-source MLLMs (i.e., GPT-5-mini, GPT-4o-mini) and open-source RGB-based MLLMs (i.e., LLaVA-Next-Video), providing a broader comparison.}
\resizebox{\textwidth}{!}{
\begin{tabular}{l>{\centering\arraybackslash}p{0.10\textwidth}|>{\centering\arraybackslash}p{0.15\textwidth}|*{2}{>{\centering\arraybackslash}p{0.07\textwidth}}|*{3}{>{\centering\arraybackslash}p{0.075\textwidth}}|*{3}{>{\centering\arraybackslash}p{0.075\textwidth}}|>{\centering\arraybackslash}p{0.085\textwidth}}
\toprule
\multirow{2}{*}{\textbf{Model}} &
\multirow{2}{*}{\textbf{Params}} &
\multirow{2}{*}{\textbf{Backbone}} &
\multicolumn{2}{c}{\textbf{Understanding}} &
\multicolumn{3}{c}{\textbf{Recognition}} &
\multicolumn{3}{c}{\textbf{Spatial Reasoning}} &
\multirow{2}{*}{\textbf{Overall}} \\
\cmidrule(lr){4-5}\cmidrule(lr){6-8}\cmidrule(l){9-11}
 &  &  & DU & CR & AR & GR & E-OCR & SR & AD & OC &  \\
\midrule
\rowcolor[HTML]{E9EAFE}
\multicolumn{12}{l}{\textbf{\textcolor{blue}{\textbullet~Closed-Source MLLMs}}} \\
GPT-5-mini &--  &--  &67.2  &67.8  &51.5  &49.2  &8.0  &42.9  &6.0  &30.2  &40.4  \\
GPT-4o-mini &--  &--  &60.5  &61.7  &50.1  &47.4  &15.5  &46.2  &22.4  &16.2  &40.0  \\
\midrule

\rowcolor[HTML]{DFF6F0}
\multicolumn{12}{l}{\textbf{\textcolor{green!50!black}{\textbullet~Open-Source MLLMs (SFT)}}} \\
LLaVA-Next-Video & 7B & Qwen2 &62.3  &62.8  &51.6  &36.2  &42.8  &41.2  &36.9  &58.2  &49.0  \\
\bottomrule
\label{tab:video_ext_results}
\end{tabular}
}
\end{table*}

\noindent\textbf{What Are the Principal Performance Bottlenecks?}
In Tab.~\ref{tab:main_results}, we report the performance of advanced closed-source MLLMs, open source MLLMs with supervised fine tuning (SFT), and event-based MLLMs across all tasks. From these results, we identify two principal bottlenecks. (1) \emph{Spatial reasoning is the main limitation in event-based MLLMs}, where even the strongest closed-source MLLMs (e.g., GPT-5), open source MLLMs (e.g., InternVL-3.5), and advanced event-based MLLMs (e.g., EventGPT+(L)) still perform poorly. (2) \emph{Spatial instruction tuning significantly improves spatial reasoning yet remains limited.} As shown in Tab.~\ref{tab:vl_sft_comparison}, spatial instruction tuning brings significant gains for open source MLLMs on spatial reasoning tasks (i.e., spatial relationship, absolute distance, and object counting), especially on object counting, suggesting that supervised fine tuning (SFT) helps reduce the cross-modal discrepancy between event streams and RGB inputs and enhances the spatial reasoning ability of MLLMs.

\noindent\textbf{Can Event SFT Mitigate the Modality Gap?} To examine the impact of supervised fine-tuning (SFT) on the performance of RGB-based MLLMs in event-based vision tasks (i.e., understanding, recognition, and spatial reasoning), we evaluate two representative open-source models (i.e., Qwen2.5-VL and InternVL-3.5) both before and after SFT, and report the results in Tab.~\ref{tab:vl_sft_comparison}. Our analysis yields two key observations. (1) \emph{SFT on event stream data contributes to reducing the modality gap.} As shown in Tab.~\ref{tab:vl_sft_comparison}, all task categories exhibit substantial performance gains after SFT, suggesting that adapting RGB-based MLLMs to event representations can effectively narrow the discrepancy between the two modalities. (2) \emph{SFT alone is insufficient to close the gap relative to natively trained event-based MLLMs.} As shown in Tab.~\ref{tab:main_results}, although the fine-tuned open-source MLLMs achieve notable improvements, they continue to fall short of event-based MLLMs trained directly on event streams. This demonstrates that the modality disparity between event stream and RGB videos remains pronounced, and that fine-tuning with million-scale event samples can mitigate the gap, though considerable disparity remains.

\section{Dynamic Event Bin Selection}
Event streams possess extremely high temporal resolution, making long-duration sequences easily exceed the context limits of MLLMs. Furthermore, the quality of motion perception depends critically on how the continuous stream is partitioned into temporal bins. To balance temporal coverage and information density under a fixed computational budget, we introduce a simple yet effective dynamic event bin selection mechanism. Instead of adopting the spatial-temporal aggregator used in EventGPT, we build a stronger baseline, termed EventGPT+, and release multiple model sizes, including EventGPT+(B) and EventGPT+(L).

To enable computationally scalable processing of the event stream, 
We consider its timestamp set 
$\mathcal{T}=\{t_1,\dots,t_M\}$ spanning the observation interval $[t_{\min}, t_{\max}]$. 
We subsequently partition this interval into $N_b$ equal-duration event bins, with the 
$k$-th bin formally defined as:
\begin{equation}
\label{eq:bin_interval}
\mathcal{I}_k = \big[\,t_{\min} + k\Delta,\; t_{\min} + (k+1)\Delta\,\big),
\end{equation}
where $\Delta = (t_{\max}-t_{\min})/N_b$.
Events assigned to $\mathcal{I}_k$ are further time-normalized by
$\tilde{t}i^{(k)} = t_i - (t{\min}+k\Delta)$, ensuring a more stable and consistent temporal alignment across varying event densities and dynamic motion patterns, thereby achieving robust and adaptive normalization with enhanced temporal coherence consistent with Algorithm~\ref{alg:dynamic_bin_selection}.

Within each bin, we build a histogram $h_k(j)$ over uniformly quantized sub-intervals and apply an overlapping sliding window of width $W$ (e.g., 10,ms with 5,ms stride). For each window start index $s$, the aggregated event count is computed as in a robust and temporally coherent manner:
\begin{equation}
C_k(s)=\sum_{j=s}^{s+W-1} h_k(j).
\end{equation}
The window with the largest event count is chosen as the representative temporal slice. Low-activity bins (where $\max_s C_k(s)$ falls below a preset threshold) fall back to their temporal center to maintain temporal coverage.

This dynamic event bin selection strategy preserves the computational budget while adaptively focusing on motion-salient segments, leading to improved temporal adaptability.


\begin{algorithm}[t]
\caption{Dynamic Event Bin Selection}
\label{alg:dynamic_bin_selection}
\begin{algorithmic}[1]
\State \textbf{Input:} Event timestamps $\mathcal{T}=\{t_1,\dots,t_M\}$; number of bins $N_b$; window size $W$
\State \textbf{Output:} Representative event subsets $\{\hat{\mathcal{E}}_k\}_{k=0}^{N_b-1}$
\State $t_{\min} \gets t_1,\; t_{\max} \gets t_M$
\State $\Delta \gets (t_{\max}-t_{\min}) / N_b$
\For{$k = 0$ to $N_b-1$}
    \State $\mathcal{I}_k \gets [\,t_{\min}+k\Delta,\; t_{\min}+(k+1)\Delta\,)$
    \State $\mathcal{E}_k \gets \{\,t_i \in \mathcal{T} \mid t_i \in \mathcal{I}_k\,\}$
    \State $\tilde{t}_i^{(k)} \gets t_i - (t_{\min}+k\Delta)$
    \State Build histogram $h_k(\cdot)$ over $\tilde{t}_i^{(k)}$
    \For{each window start $s$}
        \State $C_k(s) \gets \sum_{j=s}^{s+W-1} h_k(j)$
    \EndFor
    \State $s_k^\ast \gets \arg\max_s C_k(s)$
    \State $\hat{\mathcal{E}}_k \gets \{\,t_i \in \mathcal{E}_k \mid \tilde{t}_i^{(k)} \in [\,s_k^\ast,\; s_k^\ast+W\,)\}$
\EndFor
\State \Return $\{\hat{\mathcal{E}}_k\}_{k=0}^{N_b-1}$
\end{algorithmic}
\end{algorithm}

\begin{figure*}[t]
  \centering
   \includegraphics[width=\linewidth]{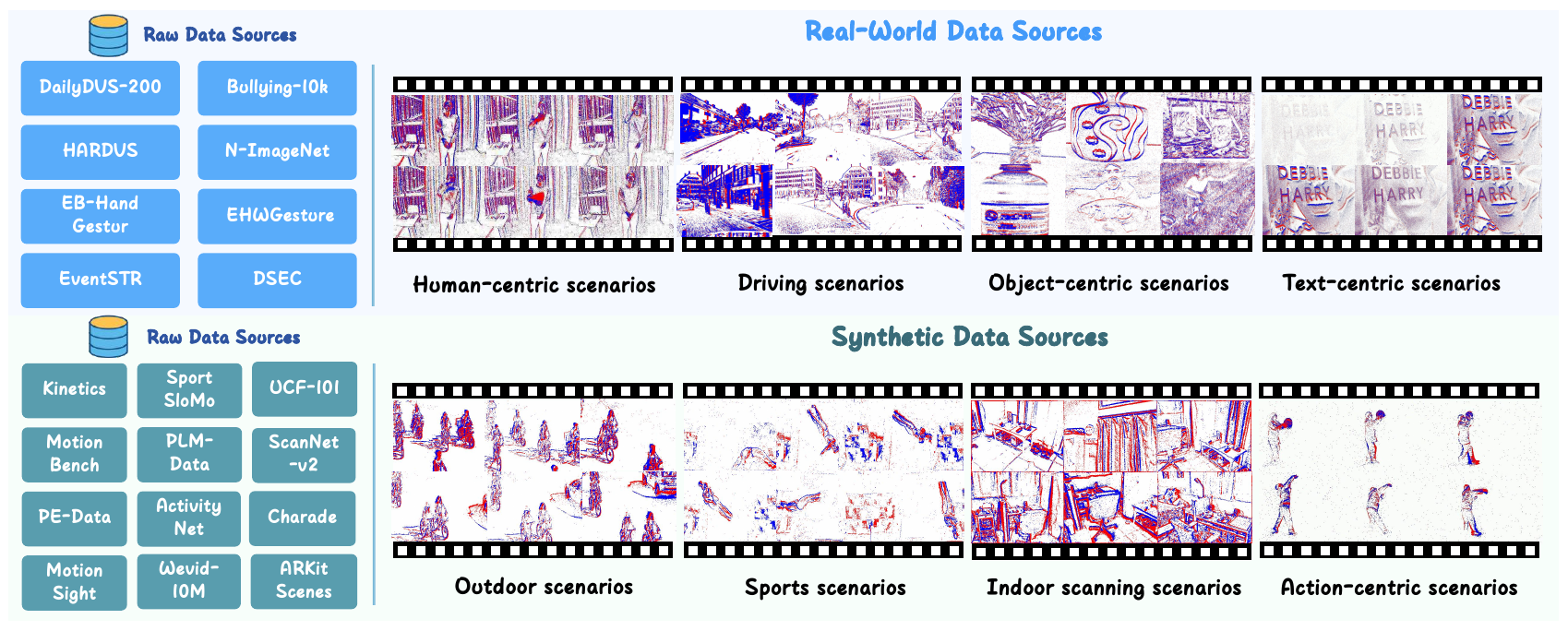}
   \caption{
Diverse data sources of EventBench encompass real-world (i.e., human-centric, driving, object-centric, text-centric) and synthetic (i.e., outdoor, sports, indoor-scanning, action-centric) scenarios. This broad coverage provides rich motion patterns, spatial structures, and semantic contexts, supporting comprehensive evaluation of event-stream understanding, recognition, and spatial reasoning.
   }
   \label{fig:data_sources}
\end{figure*}

\section{Data Source Diversity}
EventBench is constructed from a broad collection of over 20 real-world and synthetic data sources, as illustrated in Fig.~\ref{fig:data_sources}. These sources cover real-world scenarios (i.e., human-centric, driving, object-centric, and text-centric) and synthetic environments (i.e., outdoor, sports, indoor-scanning, and action-centric). This extensive scene diversity stands in sharp contrast to existing event-stream datasets and benchmarks, which are often restricted to a single domain such as driving. By encompassing heterogeneous motion patterns, spatial structures, and semantic contexts, EventBench mitigates dataset-specific overfitting and reduces evaluation bias, thereby supporting a more generalizable assessment of understanding, recognition, and spatial reasoning in event-based MLLMs.

\section{More results}
This section provides supplementary experiments, including extended evaluations over additional MLLMs in Sec.~\ref{subsec:extended_eval} and an analysis of inference efficiency under long-horizon event scenarios in Sec.~\ref{subsec:efficiency}.

\subsection{Extended Evaluation}
\label{subsec:extended_eval}
Due to space limitations in the main paper, we include additional evaluations here. To achieve a more comprehensive characterization of the capabilities of current MLLMs on event stream understanding, recognition, and spatial reasoning, we further broaden the evaluation scope. Specifically, we include additional closed-source MLLMs (i.e., GPT-5-mini, GPT-4o-mini) as well as open-source MLLMs (i.e., LLaVA-Next-Video). As shown in Tab.~\ref{tab:video_ext_results}, we report extended results covering a wider set of architectures evaluated under the same protocol. These experiments not only offer a more thorough assessment of the practical performance of existing MLLMs on event-driven tasks but also establish informative reference baselines for future research on event-based MLLMs.

\begin{table}
\centering
\small
\caption{Token-to-First-Token (TTFT) latency comparison across open-source MLLMs and open-source event-based MLLMs on the spatial reasoning subset of EventBench. }
\renewcommand{\arraystretch}{1.2}
\resizebox{\linewidth}{!}{
\begin{tabular}{l|c|c|c|c}
\toprule
\textbf{Model} & \textbf{Params}  & \textbf{LLM Backbone} & \textbf{TTFT (s)} & \textbf{Average Acc}
 \\
\midrule
\rowcolor[HTML]{DFF6F0}
\multicolumn{5}{l}{\textbf{\textcolor{green!50!black}{\textbullet~Open-Source Video MLLMs}}} \\
Qwen2.5-VL     &3B   &Qwen2.5  &2.76  &49.5 \\
InternVL-3.5   &4B   &Qwen3  &2.27  &56.2 \\
MiniCPM4-V     &4.1B   &MiniCPM4  &4.22  &52.5 \\
Mimo-VL        &7B   &MiMO  &6.78  &52.3 \\
LLaVA-Next-Video    &7B   &Qwen2  &6.89  &49.0 \\
\midrule
\rowcolor[HTML]{FEE8E4}
\multicolumn{5}{l}{\textbf{\textcolor{purple!80!black}{\textbullet~Open-Source Event-Based MLLMs}}} \\
EventGPT        &7B   &Vicuna-v1.5   &0.58   &63.5   \\
EventGPT+(B)    &2B   &Qwen2.5   &\textbf{0.49}   &63.8   \\
EventGPT+(L)    &7B   &Qwen2.5   &0.82   &\textbf{65.5}   \\
\bottomrule
\end{tabular}
}
\label{tab:ttft-comparison}
\end{table}

\subsection{Inference Efficiency Analysis}
\label{subsec:efficiency}
We evaluate TTFT on the spatial reasoning subset of EventBench, where event streams reach minute-level duration. This setting imposes the strongest temporal burden and thus best reflects long-horizon inference efficiency. The TTFT comparison conducted on an NVIDIA A800 GPU (Tab.~\ref{tab:ttft-comparison}) reveals a clear and persistent efficiency gap between RGB-based MLLMs and event-based MLLMs. Open-source RGB models (i.e., Qwen2.5-VL, InternVL-3.5, MiniCPM4-V, MiMO-VL, and LLaVA-Next-Video) exhibit high latency, ranging from \(2.27\) s to \(6.89\) s, driven by dense frame tokenization and the computational load of long video sequences. In contrast, event-based MLLMs operate within 0.49 s to \(0.82\) s, benefiting from the sparse and asynchronous structure of event streams. Under this challenging condition, EventGPT+(B) achieves the lowest TTFT at \(0.49\) s while maintaining competitive accuracy, and EventGPT+(L) delivers the highest accuracy at \(65.5\)\%, yet remains an order of magnitude faster than most RGB-based models. These results show that event-based MLLMs achieve a more favorable performance–efficiency trade-off.